\documentclass[conference, onecolumn]{IEEEtran}
\IEEEoverridecommandlockouts

\usepackage{graphicx}
\usepackage{amssymb}
\usepackage[hidelinks]{hyperref}
\usepackage{booktabs}
\usepackage[textsize=small,colorinlistoftodos,color=orange]{todonotes}
\usepackage{algpseudocode}
\usepackage{algorithm2e}
\usepackage{listings}

\newcommand\ty[1]{}
\newcommand\tx[1]{}
\title{Enhancing Traffic Incident Management with Large Language Models: A Hybrid Machine Learning Approach for Severity Classification*\\
}

\author{\IEEEauthorblockN{1\textsuperscript{st} \textbf{Artur Grigorev}}
\IEEEauthorblockA{\textit{Faculty of Engineering and IT} \\
\textit{University of Technology Sydney}\\
Sydney, Australia \\
ORCID: 0000-0001-6875-3568\\
Artur.Grigorev@uts.edu.au}

\and
\IEEEauthorblockN{2\textsuperscript{rd} \textbf{Khaled Saleh}}
\IEEEauthorblockA{\textit{Faculty of Engineering and IT} \\
\textit{University of Newcastle}\\
Newcastle, Australia \\
ORCID: 0000-0002-2589-179X\\
Khaled.Saleh@newcastle.edu.au}
\and

\IEEEauthorblockN{3\textsuperscript{th} \textbf{Yuming Ou}}
\IEEEauthorblockA{\textit{Faculty of Engineering and IT} \\
\textit{University of Technology Sydney}\\
Sydney, Australia \\
ORCID: 0000-0001-5922-9406\\
Yuming.Ou@uts.edu.au}

\and
\IEEEauthorblockN{4\textsuperscript{nd} \textbf{Adriana-Simona~Mih\u{a}i\c{t}\u{a}}}
\IEEEauthorblockA{\textit{Faculty of Engineering and IT} \\
\textit{University of Technology Sydney}\\
Sydney, Australia \\
ORCID: 0000-0001-7670-5777\\
Adriana-Simona.Mihaita@uts.edu.au}

}

\usepackage{listings}
\begin{document}
\maketitle

\section{Abstract}

This research showcases the innovative integration of Large Language Models into machine learning workflows for traffic incident management, focusing on the classification of incident severity using accident reports. By leveraging features generated by modern language models alongside conventional data extracted from incident reports, our research demonstrates improvements in the accuracy of severity classification across several machine learning algorithms. Our contributions are threefold. First, we present an extensive comparison of various machine learning models paired with multiple large language models for feature extraction, aiming to identify the optimal combinations for accurate incident severity classification. Second, we contrast traditional feature engineering pipelines with those enhanced by language models, showcasing the superiority of language-based feature engineering in processing unstructured text. Third, our study illustrates how merging baseline features from accident reports with language-based features can improve the severity classification accuracy. This comprehensive approach not only advances the field of incident management but also highlights the cross-domain application potential of our methodology, particularly in contexts requiring the prediction of event outcomes from unstructured textual data or features translated into textual representation. Specifically, our novel methodology was applied to three distinct datasets originating from the United States, the United Kingdom, and Queensland, Australia. This cross-continental application underlines the robustness of our approach, suggesting its potential for widespread adoption in improving incident management processes globally.

\hfill\break%
\noindent\textit{Keywords}: traffic accident, incident severity classification, machine learning, traffic management, large language models
\newpage

\section{Introduction}

The rise in vehicular traffic over the past few decades has led to a corresponding increase in traffic accidents, with over five million reported in the United States in 2013 alone according to the National Highway Traffic Safety Administration (NHTSA) \cite{safety2013}. This surge underscores the need for effective Traffic Incident Management Systems (TIMS) capable of handling complex datasets involving accident details, traffic conditions, and environmental factors.

A critical aspect of TIMS is the ability to classify the traffic accident severity accurately, which is essential in determining the resources required for response - including team size, equipment, and traffic control measures \cite{mary}. However, classifying accident severity poses significant challenges due to the stochastic nature of traffic accidents \cite{THEOFILATOS20163399}. Therefore, it's necessary to perform the research in the direction of finding more efficient models.

The ability of LLMs to understand and process unstructured textual data from incident reports presents a significant opportunity to augment conventional machine learning approaches. These algorithms have typically been applied to structured tabular data, but their performance can potentially be enriched by the features extracted from incident description using LLMs.

Large Language Models (LLMs), with their capability to comprehend and process unstructured textual data from accident reports (accident description), offer an opportunity to enhance the performance of traditional machine learning approaches. Traditionally, these approaches rely on structured, tabular data, which makes prediction models non-transferable due to differences in accident report formats and different structure of accident report data sets.

\textbf{Objectives:} 

1. Our primary goal is to investigate the potentials of LLMs in feature extraction from textual accident reports. By 'feature extraction', we refer to the process of selecting and encoding information from raw accident report data to represent the properties of an accident. 

\begin{figure}[h]
    \centering
    \includegraphics[width=0.7\textwidth]{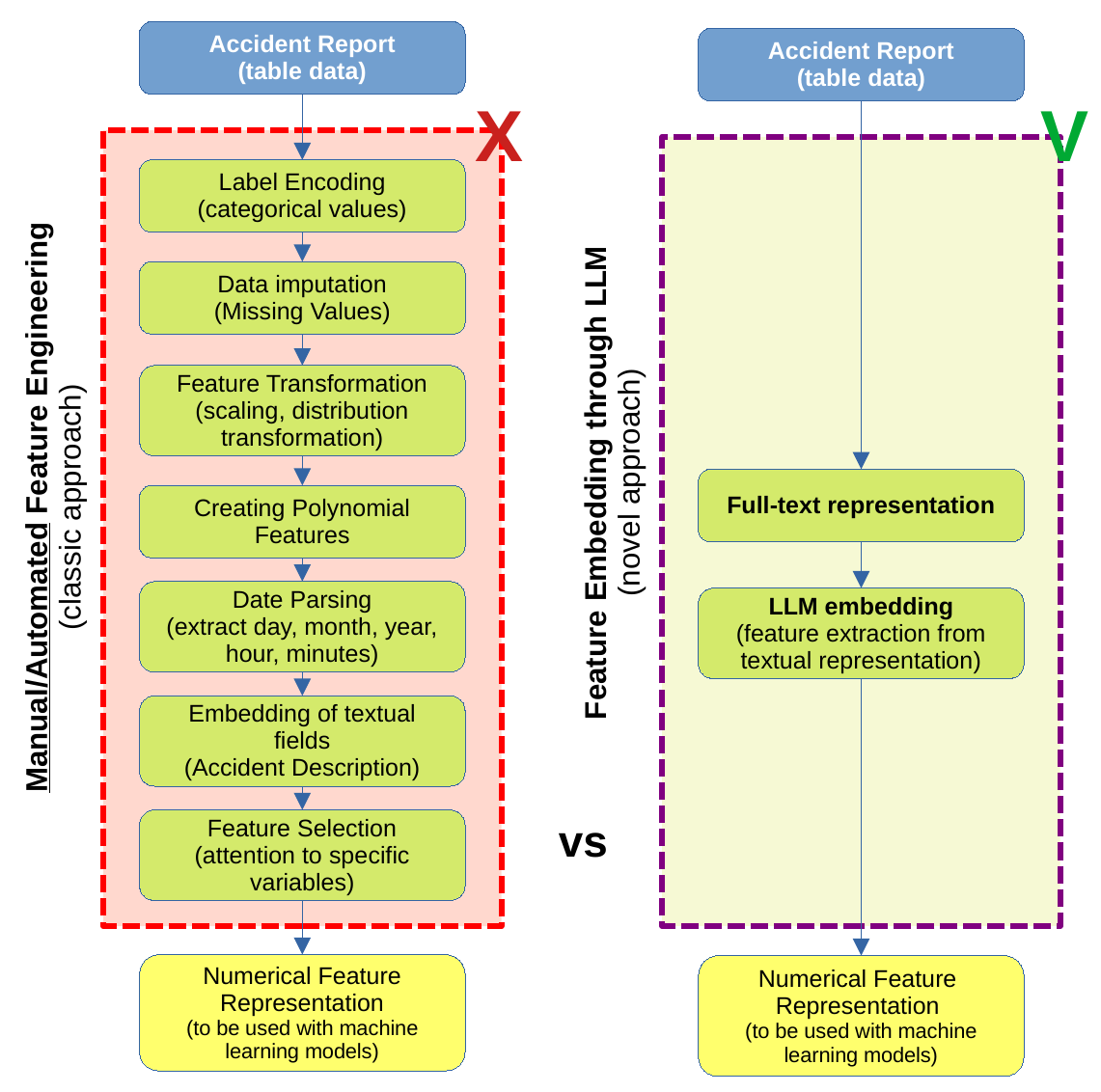}
    \caption{The advantage of using LLM models for translating unstructured text data into meaningful features for machine learning models}
    \label{fig:FT} 
\end{figure}

2. Furthermore, we aim to evaluate if the use of the extracted features, when used with traditional accident reports, can advance or at least match the performance of traditional feature engineering pipeline in classifying accident severity.  It's known that feature engineering pipeline in machine learning process has multiple complex steps, which can be mitigated by the use of full-text representation of incident reports using LLM (see Figure \ref{fig:FT}). For example, language model can be used to encode the textual representation of the date field, instead of performing complex set of steps for parsing of these values to represent them in numerical format. The use of textual accident descriptions was also found to improve the performance of traffic incident duration prediction models when incorporated with original feature set when using more simple LSTM model for feature extraction from accident narrative \cite{grigorev2022traffic}. The full-text representation allows for seamless combination of traffic accident description with other relevant accident report variables. The removal of other feature engineering steps also allows to reduce efforts in data preparation. Therefore, we seek to determine whether the use of LLM-extracted features can match manual feature engineering within traditional machine learning pipeline, and, when combined with conventional structured data, if it can  enhance the performance of traditional machine learning models (like Random Forest, XGBoost, and LightGBM) in classifying traffic incident severity.

3. The proposed methodology simplifies traditional feature engineering by using LLMs to encode full-text data representations into numerical features for machine learning models. The framework integrates LLM-extracted features with traditional machine learning techniques. Due to inherent complexity of feature engineering procedure, it may be more efficient in comparison to the use of LLM because all these steps are expected to be virtualy performed by the large language model, which creates a high demand for model's capabilities. Therefore, the difference in performance is expected. But it's unknown whether LLM approach will provide acceptable, subpar or superior performance.




\textbf{Challenges in incident management}: 
a) Solving prediction tasks in relation to traffic incidents is notoriously difficult due to their stochastic nature: accidents can happen anywhere in time and space, under various external conditions (weather, events, road closures, other cascading accidents in the network), 
b) The complexity of the task is amplified by integrating various forms of non-numerical data, such as textual incident description. Often, this data is in unstructured text form, making it more challenging to include in predictive models,
c) Various traffic departments have different ways of storing the incident data logs and the format can vary significantly from one city to another, one country to another; therefore adoption a single AI model that is efficient across multiple data sets represents a high challenge; one needs to tailor, adapt, and construct various hybrid models that can meet the challenges of each urban set-up; this can lead to a waste of computational resources. 

\textbf{Cross domain challenges and applicability of our research} The classification of traffic incident severity holds relevance and applicability across a diverse array of fields, extending well beyond the area of traffic management. This positions it as a concept with applicability to other domains.   For instance, language models were applied to the field of construction injury precursors \cite{BAKER2020103145} and construction accidents classification \cite{ZHANG2019238}. The outcomes of this research can be applied to any area of transportation, including classification of water transportation accidents \cite{yu2023water}. Also, in healthcare, this approach could be utilized to classify patient needs based on severity; in public safety, to determine the urgency of response to public complaints; or in customer service, to rank customer issues based on incoming queries. Ultimately, our goal is to streamline the data preparation process using LLMs, potentially reducing costs for traffic management organizations through time and labour savings in data-processing and improved prediction accuracy.

\textbf{Our contributions are threefold:} First, we provide an extensive comparison of combinations of various machine learning models and large language models (used for feature extraction) for the task of incident severity classification. The aim here is to identify the optimal pairing for the most accurate prediction.  Secondly, we perform a comparison of traditional feature engineering pipeline and LLM-based feature engineering processes. Thirdly, we investigate how a combination of baseline features (accident reports converted to numerical representation through the process of manual feature engineering) with LLM-extracted features can enhance the accident severity classification accuracy. Finally, our proposed method has cross-domain application potential, especially in other areas that involve predicting event outcomes based on unstructured textual data or features converted to textual representation.

\section{Related Works}

Traffic incident severity prediction is an area of extensive research, with numerous investigations employing machine learning models for accurate forecasting \cite{LiOverview2018}. These studies often process structured data sourced from Traffic Management Centres (TMC) to apply predictive models. For instance, the k-nearest neighbors algorithm and Bayesian networks have been used by some researchers to predict traffic accident severity \cite{ahmed2021comparative}. More recent studies have seen the use of ensemble methods like Random Forests and XGBoost, demonstrating their efficacy in this application \cite{Arterial2019}.

Text mining, particularly with the assistance of LLM (Language Model) such as ChatGPT, has demonstrated its effectiveness in various Natural Language Processing (NLP) tasks like summarization, information extraction, and text classification. These techniques can be applied to address different problems in the field of Intelligent Transportation Systems (ITS). One such problem is analysis of unstructured crash description, which may be crucial for documenting crash information. Also, LLm can generate a narrative summary containing details about the sequence of events, human behavior, and crash outcomes is essential. LLM models like ChatGPT prove to be efficient in generating clear, professional, and easily understandable crash summaries, assisting in evaluating the nature of the crash \cite{zheng2023chatgpt}. Another application lies in crash news mining. LLM models possess the ability to accurately extract relevant crash information from news articles and present it in a tabular form, ready for processing using machine learning algorithms.

\textbf{Limitations:} This research includes a novel approach that goes beyond existing methodologies that rely solely on machine learning models or using a single LLM model for accident severity classification \cite{agrawal2021traffic,yuan2022imbalanced,yuanlai2023text,LiOverview2018} by considering multiple Large Language Models in combination with a diverse set of predictive machine learning models. A similar multiple model approach (including both LLM and ML models) was utilized to classify accident narratives against multiple classes of outcomes \cite{goldberg2022characterizing}, but without using the entire table of accident report parameters like we do in the current research. The use of machine learning models offers various benefits of rapid training and evaluation performance (e.g. XGBoost method when using Graphical Processing Unit was shown to classify 10 million rows of 100 features each in under 45 seconds \cite{mitchell2018xgboost}), minimal memory requirements which makes possible a rich subsequent analysis including word importance estimation (e.g. using Shapley values). The key challenge in computing Shapley values lies in the number of subsets of features to be evaluated \cite{jethani2021fastshap}. For a model with $d$ features, there are $2^d$ possible subsets, since each feature can either be included or excluded from a subset. Therefore, the computational complexity for calculating the Shapley value of just one feature is exponential in the number of features, $O(2^d)$. For all $d$ features, this remains exponentially expensive. The use of fast machine learning models to process feature vectors produced by LLM makes word importance analysis at least feasible.

\textbf{Gap 1:} Overall, there appears to be a lack of comprehensive studies comparing traditional feature engineering methods with those based on LLMs. Understanding the strengths, weaknesses, and applicability of each approach in the context of traffic incident severity prediction is crucial for developing more efficient data processing pipelines and accurate predictive models. 

\textbf{Gap 2:}Also, there is a need for more extensive research to identify the most effective combinations of LLMs and ML models, since the use of LLM represents area of rapidly advancing research \cite{naveed2023comprehensive}. Such studies are essential to objectively assess the effectiveness of different methodologies and to provide clear guidance for practitioners in the field. This includes exploring different types of LLMs and ML algorithms to determine the optimal pairing for various contexts and datasets. 

\textbf{Gap 3:} Finally, there's a gap in research on methods and strategies to reduce computational complexity and enhance scalability, especially when dealing with large datasets and feature sets. For example, one of the latest data sets on traffic accidents contains 1.5 million records with 48 features each with accident narrative included \cite{moosavi2019accident}.

One of the studies that utilized advanced language models, specifically fine-tuned Bidirectional Encoder Representations from Transformers (BERT), to classify traffic injury types using a large dataset of over 750,000 crash narrative reports. The models achieved a high predictive accuracy of 84.2\% and demonstrated their effectiveness in classifying crash injury types \cite{oliaee2023using}. Previous studies using basic natural language processing (NLP) tools have limitations in handling complex sentence structures and text ambiguity, whereas advanced language models like BERT address these limitations. 

Another study regarding construction accident classification \cite{goh2017construction} explores the application of machine learning algorithms in efficiently categorizing accident narratives, using accident reports as the dataset. The researchers evaluated the performance of Support Vector Machine (SVM) \cite{hearst1998support}, K-Nearest Neighbors (KNN) \cite{peterson2009k}, Decision Tree (DT) \cite{myles2004introduction} and other machine learning algorithms on a dataset of 1000 construction accident narratives from the US OSHA website , with the use of unigram tokenization for coding accident narratives. The results showed that SVM performed the best in classifying a test set of 251 cases, with linear SVM and RBF SVM using unigram tokenization being the most effective classifiers. It signifies the importance of performance comparison of multiple models.

There are multiple studies which utilize BERT word embeddings to represent textual data from incident reports \cite{agrawal2021traffic,yuan2022imbalanced,hosseini2022application,oliaee2023using} and then utilizing various regressors to predict traffic incident duration. The BERT embeddings are commonly inputed into models like XGBoost, RandomForest, and Support-Vector models to perform the prediction task. To evaluate the effectiveness of the approach, various comparisons are made with the state-of-the-art LDA representation. LDA topic modelling is commonly used for representing textual data, but the results show that the BERT-LSTM hybrid model outperforms it in terms of mean absolute error (MAE), which indicates that the contextual understanding provided by BERT embeddings leads to more accurate predictions of traffic incident duration. The BERT model was also combined with the Recurrent Convolutional Neural Network (RCNN) model for fine-tuning \cite{yuan2022imbalanced}.

The reviewed studies showcase the diverse applications of language models, including the analysis of unstructured crash descriptions, the generation of narrative summaries, the classification of traffic seveirty and injury types. The findings indicate that the integration of advanced language models and machine learning algorithms enhances the accuracy and effectiveness of these tasks.




\section{Methodology}

Our study explores how the combination of LLM and ML models together with full-text representation can be utilized to benefit traffic accident modelling: 1) the inclusion of unstructured data from various international sources can enhance the predictive models, potentially improving the accuracy of traffic incident classification predictions in general, 2) The use of full-text representation can streamline the process of data preparation without compromising prediction accuracy.

\begin{figure*}[h]
    \centering
    \includegraphics[width=0.98\textwidth]{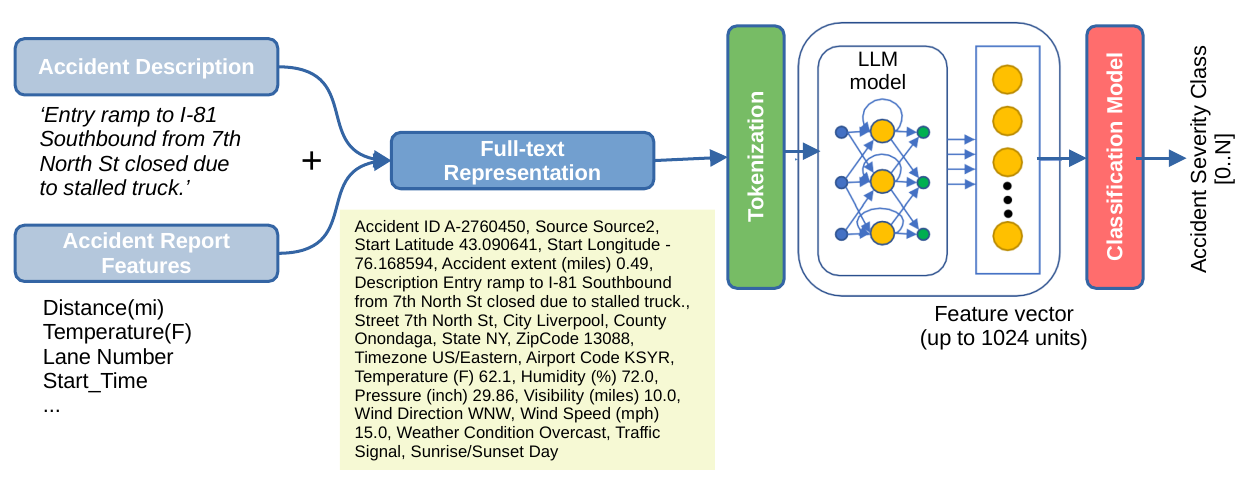}
    \caption{Diagram of the full-text representation and feature extraction}
    \label{fig:D} 
\end{figure*}

The full-text representation is an approach to represent both tabular data from accident reports and accident narrative into a single text string which then can be used to perform feature extraction using various LLM models (see Figure \ref{fig:D}): accident report values are combined with their corresponding names including the accident narrative. Tokenization is performed before the feature extraction - it is a fundamental step that involves dividing text into smaller units, known as tokens, to facilitate processing of the text by the LLM model. These tokens can be words (e.g. numbers or abbreviations) or parts of words (e.g. parts of a highway index: I-81 is being split into 'I', '-' and '81').

These extracted features (numerical representations of textual accident reports) can then be used with machine learning models to perform traffic accident classification. As shown previously (see Figure \ref{fig:FT}), we expect LLM models to be able to intrinsicaly perform various tasks of data representation including label encoding, creating polynomial features (since reliance on convolutional layers in architecture), handling missing data (offered by intrinsic flexibility of text representation), performing feature selection (by utilizing attention mechanism) and keyword extraction from narrative (frequently utilized approach in machine learning to create binary variables based on word presence in the text \cite{qader2019overview}).

BERT (Bidirectional Encoder Representations from Transformers) \cite{devlin2018bert} utilizes a training method known as Masked Language Modeling (MLM). Its notable features include the introduction of bidirectional context, allowing the model to better understand the semantics of each word. Additionally, BERT introduced a distinct pretraining and fine-tuning methodology that has been widely used for incident analysis \cite{goh2017construction,agrawal2021traffic,yuan2022imbalanced}.

BERT-large \cite{devlin2018bert} also employs Masked Language Modeling (MLM) as its training method. As an enlarged version of BERT, it maintains the core principles but scales up the architecture to handle more complex tasks and deliver better performance.

XLNet \cite{yang2019xlnet} uses a Generalized Autoregressive Pretraining method. It addresses some limitations of BERT and incorporates features of Transformer-XL, thus allowing for better handling of long-term dependencies in the text.

XLNet-large \cite{yang2019xlnet} is a scaled-up version of XLNet, and it also employs Generalized Autoregressive Pretraining. It brings the advantages of XLNet into a larger and more powerful architecture, suitable for more complicated tasks.


RoBERTa (Robustly Optimized BERT) \cite{liu2019roberta} aims to improve upon BERT by optimizing its training process. It employs a modified form of MLM as its training method, removes the next sentence prediction task, and introduces dynamic masking for better performance.

ALBERT (``A Lite BERT'') \cite{lan2019albert} is a self-supervised transformer-based NLP model that utilizes Masked Language Modeling and Sentence Ordering Prediction for pretraining, utilizes shared-layer architecture for reduced memory footprint. The latest version (2.0) improves upon its predecessor with adjustments like lower dropout rates and additional training data for enhanced performance in downstream tasks. The largest configuration is ALBERT-xxlarge version 2 have been selected as a large model in this study.

Including variations of BERT in the analysis is a design choice due to the expectation of higher performance from these models. BERT and its variants are designed to capture complex language nuances and contextual relationships within text, which often result in better performance on tasks involving natural language understanding. Even if current datasets do not show substantial differences between language models, it remains a reasonable approach to test BERT variations, as they have the potential to yield superior results, especially with richer narrative content in the data.

The model summary table (see Table \ref{table:comprehensive_nlp_models}) provides a synthesized overview of several prominent NLP models,  the technical aspects and functionalities of these models, number of parameters, vector size, primary features, and their relevance to traffic incident severity classification. This comparative analysis helps to understand the capabilities of each model when dealing with incident report texts, thereby informing our selection of models for the study. The descriptions of each model's characteristics highlight their suitability for processing and analyzing text from traffic incident reports. 

\begin{table*}[h]
\centering
\small
\begin{tabular}{|p{2.0cm}|p{1.8cm}|p{1.5cm}|p{6.0cm}|p{4.5cm}|}
\hline
\textbf{Model (Reference)} & \textbf{Num. of Parameters} & \textbf{Vector Size} & \textbf{Notable Features} & \textbf{Relevance for the Study} \\
\hline
BERT \cite{devlin2018bert} & 110 mil & 768 & Masked Language Modeling (MLM), Bidirectional context, Pretrain-finetune discrepancy. Wide application in similar NLP tasks and versatility in fine-tuning. & Effective contextual understanding of text. Widely used for incident analysis. \\
\hline
BERT-large \cite{devlin2018bert} & 345 mil & 1024 & Masked Language Modeling (MLM), Bidirectional context, Pretrain-finetune discrepancy. Capability to process complex tasks and large text sequences. Expected higher performance due to larger architecture. & Enlarged version with improved contextual capturing. \\
\hline
XLNet \cite{yang2019xlnet} & 110 mil & 768 & Generalized Autoregressive Pretraining, Overcomes BERT limitations, Transformer-XL integration. Superior handling of sequence prediction and permutation-based training. & Addresses BERT's limitations and incorporates long-term dependencies in text. \\
\hline
XLNet-large \cite{yang2019xlnet} & 340 mil & 1024 & Generalized Autoregressive Pretraining, Overcomes BERT limitations, Transformer-XL integration. Capability to process complex tasks and large text sequences. Expected higher performance due to larger architecture. & Enlarged version with improved contextual capturing. \\
\hline
RoBERTa \cite{liu2019roberta} & 125 mil & 768 & Optimized BERT (MLM with changes), Longer training, Removed next sentence prediction, Dynamic masking. Better performance due to optimized training and handling of more data during pretraining. & Improved training process for BERT-like models. \\
\hline
RoBERTa-large \cite{liu2019roberta} & 355 mil & 1024 & Optimized BERT (MLM with changes), Longer training, Removed next sentence prediction, Dynamic masking. Capability to process complex tasks and large text sequences. Expected higher performance due to larger architecture. & Enlarged version with improved contextual capturing. \\
\hline
ALBERT \cite{lan2019albert} & 18.2 mil & 768 & Optimized BERT (MLM with changes), Sentence Ordering Prediction, Layer-Sharing Architecture, Reduced Memory Footprint. Implementation efficiencies and advancements over BERT with comparable abilities in context understanding. & Reduces BERT's memory footprint with shared-layer architecture and efficient pretraining tasks. \\
\hline
ALBERT-large \cite{lan2019albert} & 223 mil & 4096 (ALBERT-xxlarge) & Optimized BERT (MLM with changes), Sentence Ordering Prediction, Layer-Sharing Architecture, Reduced Memory Footprint. Capability to process complex tasks and large text sequences. Expected higher performance due to larger architecture. & Enlarged version with improved contextual capturing. \\
\hline
\end{tabular}
\caption{Comprehensive Overview of NLP Models, their Features, and Relevance}
\label{table:comprehensive_nlp_models}
\end{table*}


We also perform text traffic accident severity classification using various machine learning models:

XGBoost \cite{chen2016xgboost} is an advanced gradient boosting algorithm that utilizes the principle of boosting weak learners using the gradient descent architecture. It features several advanced techniques such as regularization (L1 and L2), which prevents overfitting and improves model generalization. XGBoost also supports various objective functions including regression, classification, and feature ranking.

LightGBM \cite{ke2017lightgbm}, standing for Light Gradient Boosting Machine, is a gradient boosting framework that uses tree-based learning algorithms. Its main advantage lies in its use of Gradient-based One-Side Sampling (GOSS) and Exclusive Feature Bundling (EFB), which collectively reduce the data size and feature space without compromising accuracy. LightGBM is designed for distributed and efficient training, particularly on large datasets, and supports categorical features natively.

RandomForest (RF) \cite{liaw2002classification} aggregates the predictions of multiple decision trees to improve predictive accuracy and control overfitting. Each tree in the RandomForest is built from a sample drawn with replacement (bootstrap sample) from the training set. Moreover, when splitting a node during the construction of a tree, the split that is chosen is no longer the best split among all features. Instead, the split that is picked is the best split among a random subset of the features. This results in a wide diversity that generally results in a better model.

K-Nearest Neighbors (KNN) \cite{peterson2009k} operates by finding the predefined number \( k \) of training samples closest in distance to the new point and predict the label from these. The distance can be any metric measure: standard Euclidean distance is the most common choice. KNN has the advantage of being simple to interpret and having little to no assumption about the data.

Machine learning models have various specifics in application and require performance considerations (see Table \ref{tab:model_comparisons}).

\begin{table}[h]
\centering
\begin{tabular}{|p{0.2\textwidth}|p{0.2\textwidth}|p{0.2\textwidth}|p{0.2\textwidth}|}
\hline
\multicolumn{1}{|c|}{\textbf{Model}} & \multicolumn{1}{|c|}{\textbf{Description}} & \multicolumn{1}{|c|}{\textbf{Specifics}} & \multicolumn{1}{|c|}{\textbf{Performance}} \\
\hline
XGBoost & Gradient boosting utilizing L1/L2 regularization & Exhaustive gradient-based approach & Fast training with exhaustive search; iteratively refines model \\
\hline
LightGBM & Gradient boosting employing GOSS and EFB & Approximate gradient-based approach & Faster than XGBoost, ideal for large datasets \\
\hline
RandomForest & Ensemble of decision trees built on bootstrapped samples & Random feature/rows subset selection for training and averaging of predictions & Robust but relatively slower than gradient boosting methods \\
\hline
K-Nearest Neighbors (KNN) & Instance-based method using distance metrics & Performance significantly depends on hyper-parameters - distance metric and k need to be appropriately chosen & Can be computationally expensive but easy interpretation via neighborhood concept \\
\hline
\end{tabular}
\caption{Comparison of selected machine learning models}
\label{tab:model_comparisons}
\end{table}

Overall, XGBoost and RandomForest can be computationally expensive when used with high-dimensional vectors provided by language models. KNN performance significantly decreases with increasing number of samples. There is a need for computationally effective methods that can be used with LLM features to perform classification and regression tasks. LightGBM may provide an effective solution for these tasks. Moreover, the use of tree-based methods can be more practical due to low dependence on hyper-parameters, in comparison to artificial neural networks where number of units, types of units, number of layers and training method have a high impact on final result, which requires a lot of trial-and-error attempts in network design \cite{OZESMI200683, ren2021comprehensive}. Tree-based methods provide a streamlined approach to utilizing language model features.

\section{Case study \& Experiment Setup}

The three datasets, originating from the United States, the United Kingdom, and Queensland (Australia), offer varied but rich contextual information about traffic accidents (see Table \ref{fig:T}). The U.S. dataset consists of 31 fields (reduced from 49 due to zero variance in columns) and has emphasis on environmental and lighting conditions, such as weather and astronomical twilight state, which could be crucial for understanding the impact of these factors on accidents. It also includes latitude and longitude for both the start and end points of an accident (as sourced from MapQuest and Bing services), as well as a detailed description of the accident itself. In contrast, the UK dataset comprises 34 fields with a focus on infrastructure details like pedestrian crossings and local authority information, which can be valuable for urban planning and public policy analyses. The Queensland dataset is the most comprehensive with 41 fields, providing hyper-local geographical information, including the suburb and local government area, as well as specific roadway features and traffic control setups. This makes the Queensland data potentially useful for localized analysis. While all three datasets contain some form of time and location information, they vary significantly in the types of fields and the level of detail, suggesting that accident report unification may be required for the transferability of any international accident model. Overall, each dataset has its own unique focus—environmental factors and accident extent in the U.S., infrastructure in the UK, and detailed geographical and situational context in Queensland.

According to methodology, the table representation of accident reports is converted into a full text representation - all the columns simply converted to text in the form ``columns name: column value'' in a single string of text.

Large Language Models are primarily optimized for natural language and may find it challenging to accurately interpret short strings full of domain-specific abbreviations and numbers unless specifically fine-tuned for such tasks.


\begin{table}[h!]
\centering
\begin{tabular}{|p{0.3\textwidth}|p{0.65\textwidth}|}
\hline
\textbf{Description} & \textbf{Example Crash Report}\\
\hline
USA Data Set & Accident ID A-7463401, Source Source1, Start Latitude 32.68116, Start Longitude -97.02426, End Latitude 32.67618, End Longitude -97.03483, Accident extent (miles) 0.704, Description Ramp to I-20 Westbound - Accident., Street President George Bush Tpke S, City Grand Prairie, County Dallas, State TX, ZipCode 75052, Timezone US/Central, Airport Code KGPM, Temperature (F) 48.2, Humidity (\%) 75.0, Pressure (inch) 30.26, Visibility (miles) 10.0, Wind Direction South, Wind Speed (mph) 5.8, Weather Condition Mostly Cloudy, Junction, Sunrise/Sunset Night, Civil Twilight Night, Nautical Twilight Night, Astronomical Twilight Night, Start\_Time\_hour 22, Start\_Time\_month 1, Weather\_Timestamp\_hour 22, Weather\_Timestamp\_month 11\\
\hline
UK Data Set & accident\_index: 2018460317259, accident\_year: 2018, accident\_reference: 460317259, location\_easting\_osgr: 556147.0, location\_northing\_osgr: 165830.0, longitude: 0.241871, latitude: 51.370065, police\_force: 46, number\_of\_vehicles: 1, number\_of\_casualties: 1, date: 08/08/2018, day\_of\_week: 4, time: 11:35, local\_authority\_district: 538, local\_authority\_ons\_district: E07000111, local\_authority\_highway: E10000016, first\_road\_class: 3, first\_road\_number: 20, road\_type: 6, speed\_limit: 60, junction\_detail: 3, junction\_control: 4, second\_road\_class: 6, second\_road\_number: 0, pedestrian\_crossing\_human\_control: 0, pedestrian\_crossing\_physical\_facilities: 0, light\_conditions: 1, weather\_conditions: 1, road\_surface\_conditions: 1, special\_conditions\_at\_site: 0, carriageway\_hazards: 0, urban\_or\_rural\_area: 2, did\_police\_officer\_attend\_scene\_of\_accident: 1, trunk\_road\_flag: 2, lsoa\_of\_accident\_location: E01024433\\
\hline
Queensland (Australia) Data Set & Crash\_Ref\_Number: 28863.0, Crash\_Year: 2004.0, Crash\_Month: September, Crash\_Day\_Of\_Week: Wednesday, Crash\_Hour: 6.0, Crash\_Nature: Angle, Crash\_Type: Multi-Vehicle, Crash\_Longitude: 152.872284325108, Crash\_Latitude: -27.5455985592659, Crash\_Street: Kangaroo Gully Rd, Crash\_Street\_Intersecting: Mount Crosby Rd, State\_Road\_Name: Mount Crosby Road, Loc\_Suburb: Anstead, Loc\_Local\_Government\_Area: Brisbane City, Loc\_Post\_Code: 4070, Loc\_Police\_Division: Indooroopilly, Loc\_Police\_District: North Brisbane, Loc\_Police\_Region: Brisbane, Loc\_Queensland\_Transport\_Region: SEQ North, Loc\_Main\_Roads\_Region: Metropolitan, Loc\_ABS\_Statistical\_Area\_2: Pinjarra Hills - Pullenvale, Loc\_ABS\_Statistical\_Area\_3: Kenmore - Brookfield - Moggill, Loc\_ABS\_Statistical\_Area\_4: Brisbane - West, Loc\_ABS\_Remoteness: Major Cities, Loc\_State\_Electorate: Moggill, Loc\_Federal\_Electorate: Ryan, Crash\_Controlling\_Authority: State-controlled, Crash\_Roadway\_Feature: Intersection - T-Junction, Crash\_Traffic\_Control: No traffic control, Crash\_Speed\_Limit: 70 km/h, Crash\_Road\_Surface\_Condition: Sealed - Dry, Crash\_Atmospheric\_Condition: Clear, Crash\_Lighting\_Condition: Daylight, Crash\_Road\_Horiz\_Align: Curved - view open, Crash\_Road\_Vert\_Align: Level, Crash\_DCA\_Code: 202.0, Crash\_DCA\_Description: Veh'S Opposite Approach: Thru-Right, Crash\_DCA\_Group\_Description: Opposing vehicles turning, DCA\_Key\_Approach\_Dir: E, Count\_Unit\_Car: 1.0, Count\_Unit\_Motorcycle\_Moped: 1.0, Count\_Unit\_Truck: 0.0, Count\_Unit\_Bus: 0.0, Count\_Unit\_Bicycle: 0.0, Count\_Unit\_Pedestrian: 0.0, Count\_Unit\_Other: 0.0\\
\hline
\end{tabular}
\caption{Example of full text representations for different data sets}
\label{fig:T}
\end{table}

\subsection{Data preparation}
The preprocessing of data is a crucial step in developing our model. Initially, we have an imbalance in the sample distribution across different severity classes. To resolve this, we use an even sampling. This function balances the dataset by sampling even number of samples (e.g. 12,500 for USA data set) instances from each unique class present in the 'Severity' column. This even sampling ensures a fair representation of each severity level in the data we analyze. Secondly, we identify and remove zero-variance features from the dataset. These are columns that contain only a single unique value and are not beneficial for the modeling process. In fact, some algorithms require the removal of zero-variance features for successful convergence. Finally, we enhance the feature set by processing the time-based information where it's available. Specifically, we calculate and include the starting hour and the month of each accident. This addition aims to capture any temporal patterns that might exist in the occurrence of accidents.

The table \ref{table:experiment_matrix} outlines three scenarios focused on the task of Severity Classification of traffic accidents using LLM. 

\begin{itemize}
\item Severity Classification with Baseline Accident Report Features: In this scenario, the model is trained using only numerical baseline accident report features obtained from tabular representation of accident reports. Objective: To assess how well traditional, structured data performs in predicting the severity of traffic accidents.
\item  Severity Classification with NLP Features:  This scenario focuses on training the model using features derived from fulltext representation and LLM. Objective: To determine the performance of the proposed fulltext+LLM approach.
\item  Severity Classification with a Combination of Baseline and NLP Features: This scenario combines both baseline accident report features and NLP-derived features for model training. Objective: To evaluate if a combination of structured and unstructured data improves the model’s ability to predict the severity of an accident. All the features used in scenarios 1 and 2 are combined into a single feature set.
\end{itemize}


\begin{table*}[h]
\centering
\small
\renewcommand{\arraystretch}{1.5}
\begin{tabular}{c|cccc}
\toprule
& \multicolumn{4}{c}{\textbf{Severity Classification}} \\
\midrule
& \textbf{LightGBM} & \textbf{Random Forest} & \textbf{XGBoost} & \textbf{KNN} \\
\midrule
\textbf{NLP Features (BERT, etc)} & 
\begin{tabular}[c]{@{}c@{}}\checkmark\end{tabular} & 
\begin{tabular}[c]{@{}c@{}} \checkmark\end{tabular} & 
\begin{tabular}[c]{@{}c@{}} \checkmark\end{tabular} & 
\begin{tabular}[c]{@{}c@{}}\checkmark\end{tabular} \\
\midrule
\textbf{Baseline Features (report)} & 
\begin{tabular}[c]{@{}c@{}} \checkmark\end{tabular} & 
\begin{tabular}[c]{@{}c@{}}\checkmark\end{tabular} & 
\begin{tabular}[c]{@{}c@{}} \checkmark\end{tabular} & 
\begin{tabular}[c]{@{}c@{}} \checkmark\end{tabular} \\
\midrule
\textbf{Combined Features\newline (report+NLP)} & 
\begin{tabular}[c]{@{}c@{}}\checkmark\end{tabular} & 
\begin{tabular}[c]{@{}c@{}}\checkmark\end{tabular} & 
\begin{tabular}[c]{@{}c@{}} \checkmark\end{tabular} & 
\begin{tabular}[c]{@{}c@{}} \checkmark\end{tabular} \\
\bottomrule
\end{tabular}
\caption{Extended Matrix of Experiments for Severity Classification Using Various Models}
\label{table:experiment_matrix}
\end{table*}

%

%
%
%

\section{Results}

Next figures demonstrate the accident severity classification performance over different feature sets (report only features, NLP - features extracted from the full text representation of accident reports as discussed previously, report+NLP - combination of report only features with extracted features from a language model).
Results represent the cross-validation results of different models and feature sets. We assess the models based on 4 metrics: Accuracy, F1-score, Precision, Recall.

The findings are as follows:
\begin{itemize}
\item F1-score is representative for the evaluation of the models and may be prioritized for interpretation and further evaluations over accuracy, precision and recall.
\item There is no apparent difference between all the language models. This could be attributed to the limited narrative content within the accident reports, suggesting that the models are primarily leveraging tabular data. If the textual data does not contain much discriminative information (e.g. complex textual descriptions), language models may not have sufficient context or features to significantly outperform each other, leading to a uniformity in their performance metrics.
\item Also, there is a negligible difference between base and large variants of models (e.g. bert vs bert-large). Accident reports are represented as small paragraphs of text, where advantages of language models in interpreting language features may not be relevant.
\item The ability to use text representation right away, while achieving accceptable prediction performance, instead of feature engineering (e.g. normalization of values, label encoding, functional feature transformations, date interpretation, etc) is of interrest to traffic management authorities and data analysts in transportation.
\item In general, the use of additional language features together with report features may improve the severity classification performance. It may lead to the use of vectors of higher dimensionality, but tree ensemble models show sufficient performance in this case.
\end{itemize}


These results suggest that machine learning models trained with a combination of LLM-features and report-features tend to perform better at predicting traffic incident severity class. The combination of BERT and XGBoost specifically presents a robust method in this task.

\begin{figure*}[h]
    \centering
    \includegraphics[width=0.85\textwidth]{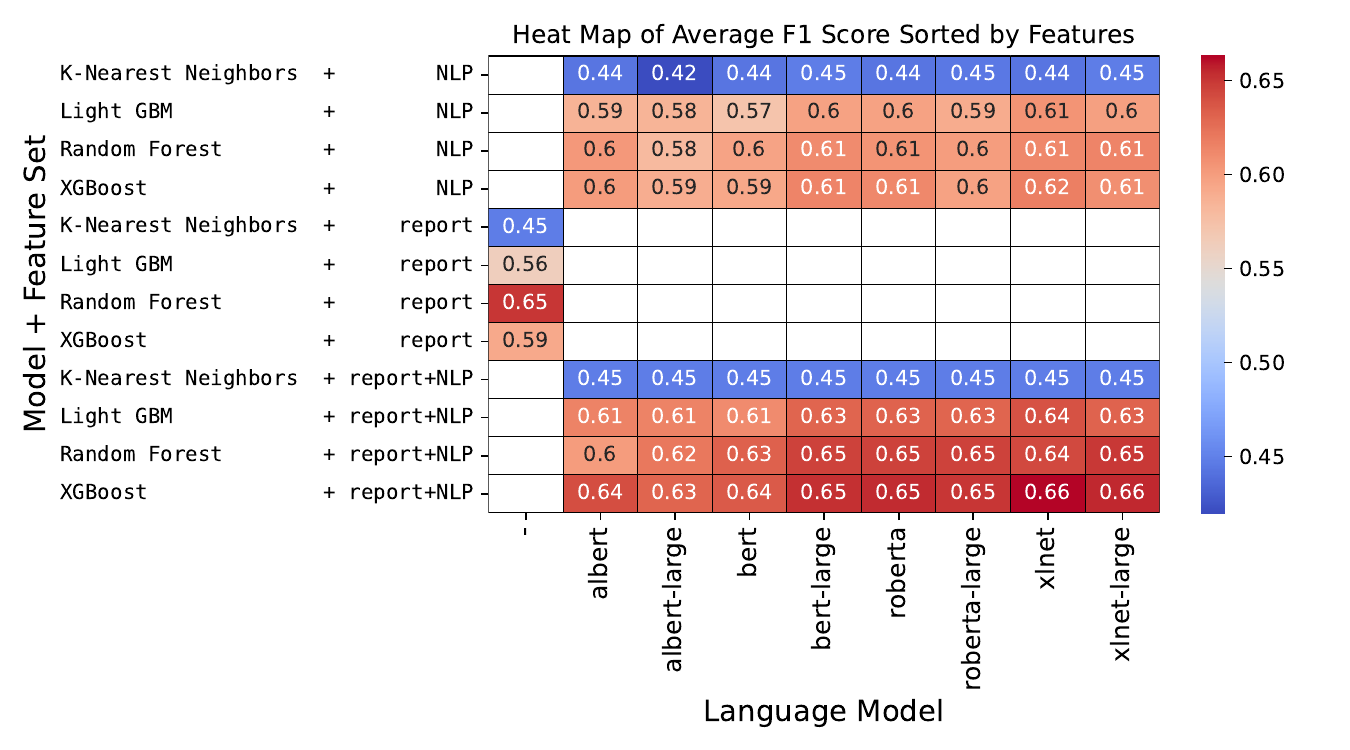}
    \caption{Random Forest for Queensland, Australia}
    \label{fig:Q-F1} 
\end{figure*}

For Queensland data set, the best performing model is the combination of report and language features (extracted using GPT-2) with RandomForest reaching F1-score of 0.65 (see Figure \ref{fig:Q-F1}). 

\begin{figure*}[h]
    \centering
    \includegraphics[width=0.85\textwidth]{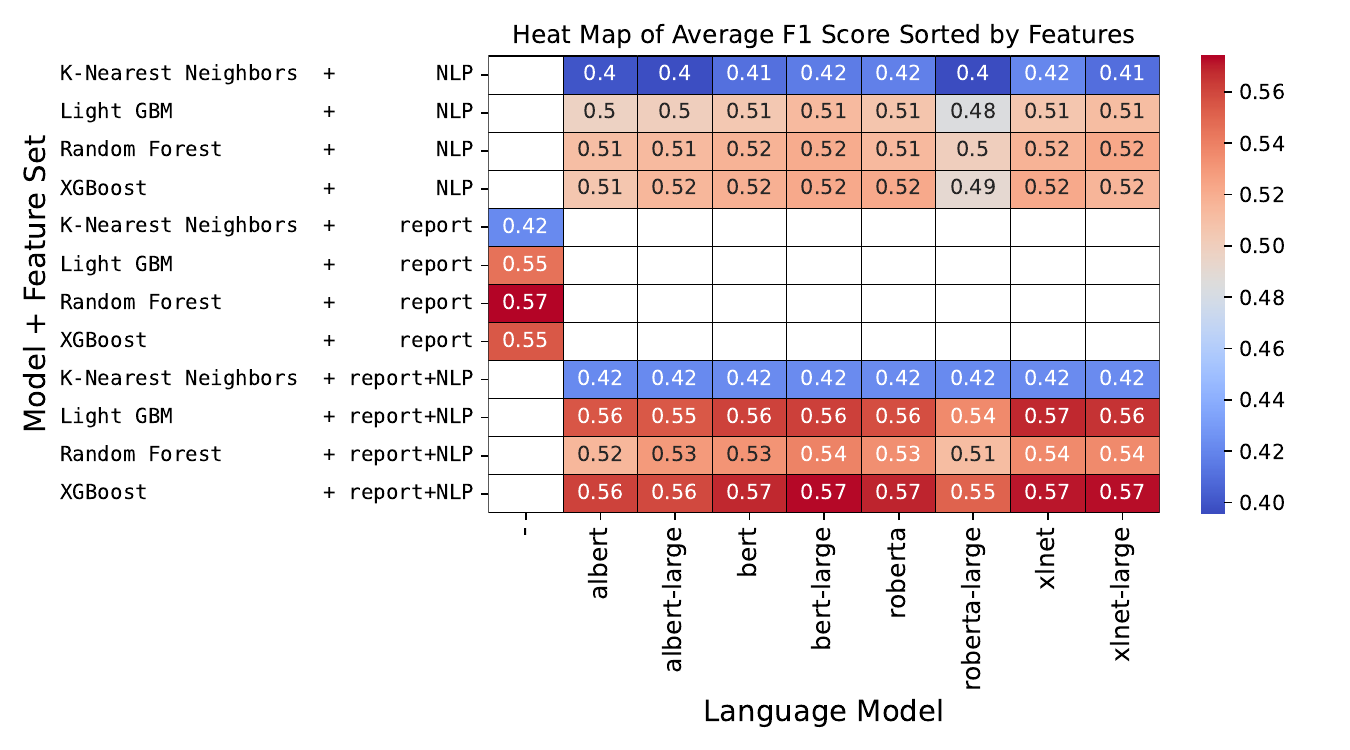}
    \caption{Random Forest for UK}
    \label{fig:UK-F1} 
\end{figure*}

In our analysis, it is evident that certain combinations of features and models outperform others. Notably, the highest F1-score of 0.58 was achieved by a RandomForest model. What makes this particular model stand out is its use of both report and language features, the latter being extracted via the GPT-2 language model, as detailed in Figure \ref{fig:UK-F1}. Following closely is the XGBoost model integrated with BERT features, which reached an F1-score of 0.56, also displayed in Figure \ref{fig:UK-F1}. When it comes to language models, the results did not show any significant performance variation between different language models.

\begin{figure*}[h]
    \centering
    \includegraphics[width=0.85\textwidth]{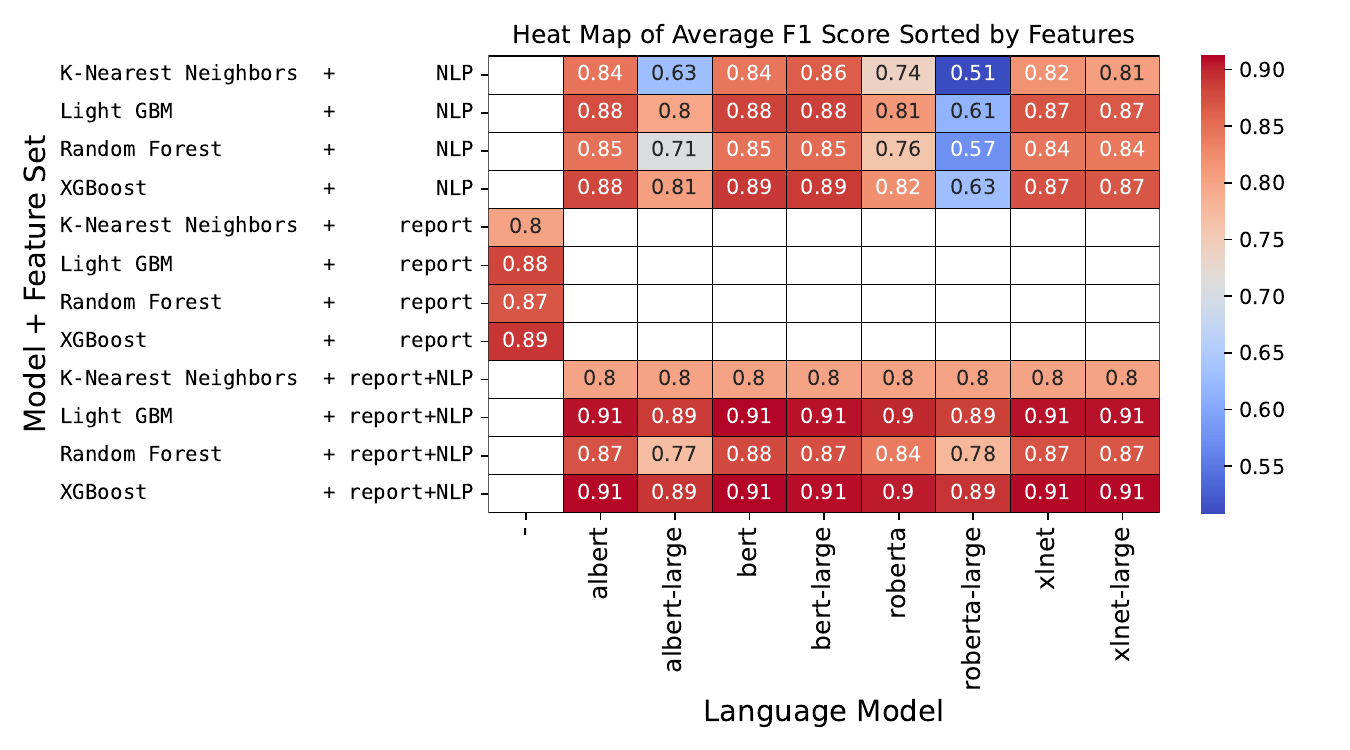}

    \caption{Random Forest for USA}
    \label{fig:USA-F1} 
\end{figure*}

The USA data set has rather unique response to the use of language models in comparison to two other data sets (Queensland and UK), see Figure \ref{fig:USA-F1}

\begin{itemize}
\item The best performing model is the combination of report and language features (extracted using BERT) with XGBoost reaching F1-score of 0.89. In comparison to using report features only - F1-score is 0.82.
\item The performance of using NLP features from different models can be ordered in the following way: 1) BERT, 2) BERT-large, 3) XLNet-large, 4) XLNet, 5) Roberta. 
\item What is highly important is that BERT model can be effectively used with full text representation of accident reports right away and demonstrate performance higher than just using accident reports.
\item XGBoost and RandomForest show the best performance and may be used based on model preference.
\item The rest of the models have basic architecture and perform worse than tree ensembles. These models still demonstrate low but unusual performance - using only language features with LogisticRegression or KNN shows much better performance (0.81 and 0.84 correspondingly) than just using accident reports (F1-score is 0.6). These results highlight the applicability of combining older or simpler models with advanced language models.
\end{itemize}

In general, the performance of RandomForest is highest, with XGboost as second best model. In all the cases, the performance of combining language feature sets with accident report features shows lower performance than just using accident report features.

For Uk data set, the performance of RandomForest is highest with negligible variation betwwen language models, with XGboost as second best model. In all the cases, the performance of combining language feature sets with accident report features shows lower performance than just using accident report features. When using features extracted using GPT-2 the performance is the lowest when using each of ML models.



The analysis of the USA dataset reveals that XGBoost outperforms other models when utilizing either report-only features or a combination of report and language features. The bert-large variant demonstrates a marginally better performance compared to the base BERT model. All other models have lower performance metrics than the BERT. So far, this data set shows both the highest severity classification performance and advancement of using language model features.

\begin{figure*}[h]
    \centering
    \includegraphics[width=0.95\textwidth]{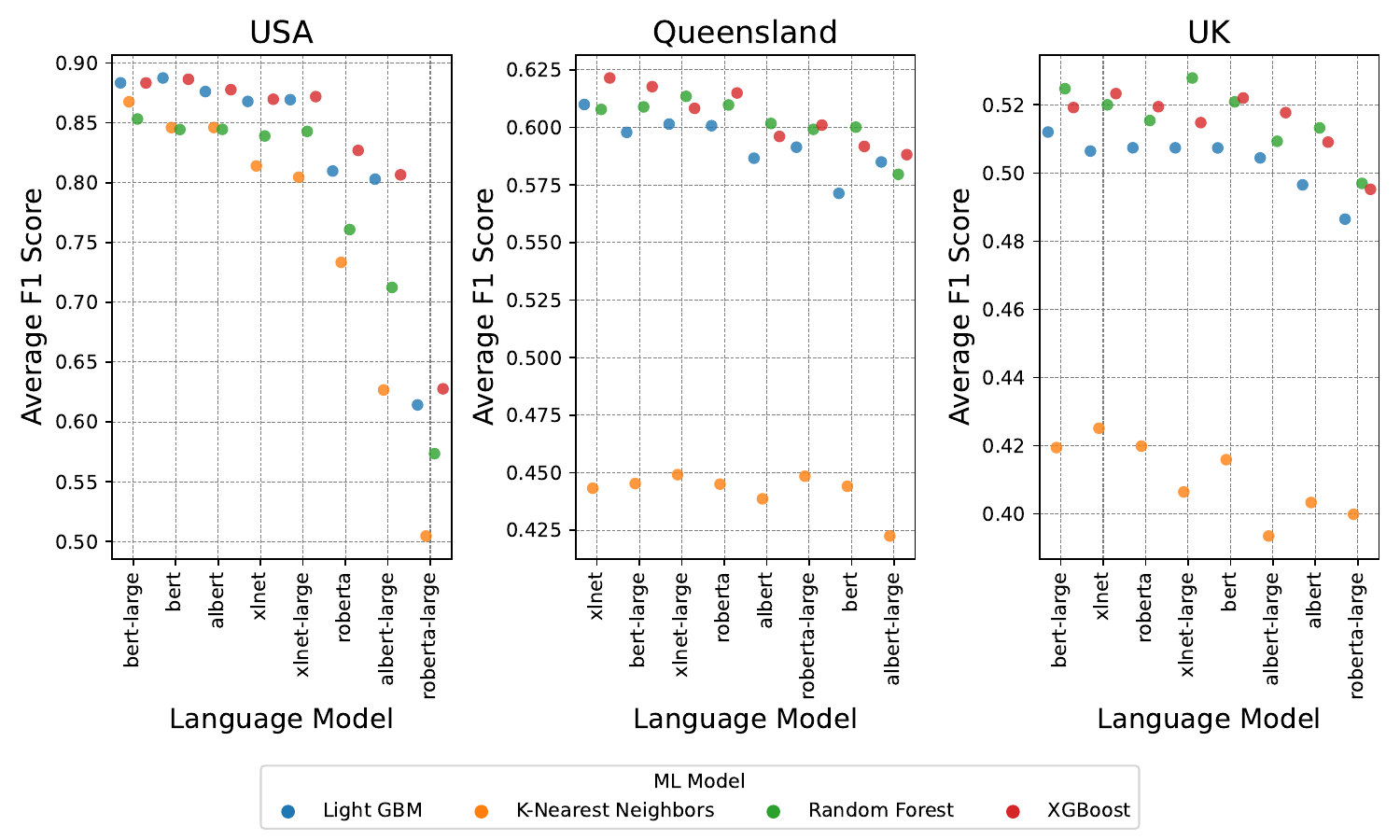}
    \caption{General comparison of Average F1 Score for traffic accident severity classification on USA data set using incident description only}
    \label{fig:jitter} 
\end{figure*}

The performance of different language models was further analyzed through a point plot, with the models arranged on the x-axis in descending order of their average F1 scores (see Figure \ref{fig:jitter}). Across all three datasets, roberta-large and albert-large show lowest performance, while bert-large and xlnet models consistently show high performance levels. XGBoost displayed competitive results in comparison to RandomForest, with only negligible differences in their performance. The K-Nearest Neighbors algorithm was observed to have the lowest performance.

\subsection{Performance comparison}

The computational experiments were conducted on a high-performance computing system equipped with dual Intel Xeon Gold 6238R CPUs, each featuring 28 cores operating at a base frequency of 2.20 GHz, capable of reaching up to 4.00 GHz. The system follows a 64-bit architecture, supports multi-threading with a total of 56 logical CPUs, and is optimized for complex computational tasks through advanced virtualization (VT-x), and large-scale memory operations. The system is further enhanced with NVIDIA Quadro RTX 6000 graphics card, equipped with 24 GB of GDDR6 memory. 

\begin{figure}[h]
    \centering
    \includegraphics[width=0.4\textwidth]{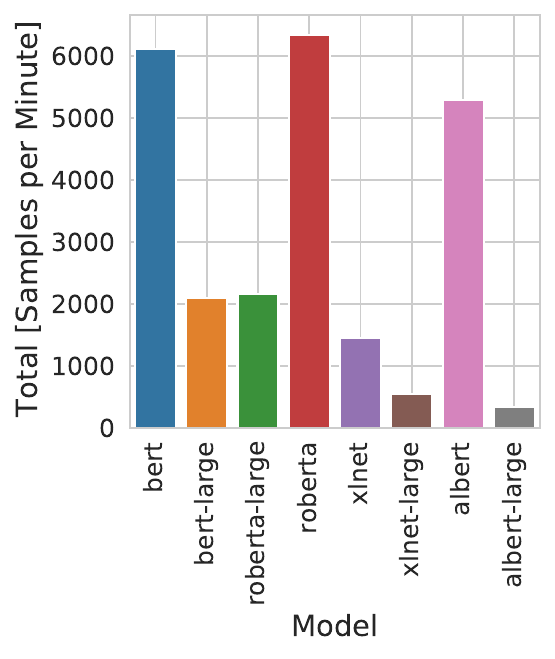}
    \caption{Language model performance (total): samples per minute}
    \label{fig:totalperff1} 
\end{figure}

\begin{figure}[h]
    \centering
    \includegraphics[width=0.4\textwidth]{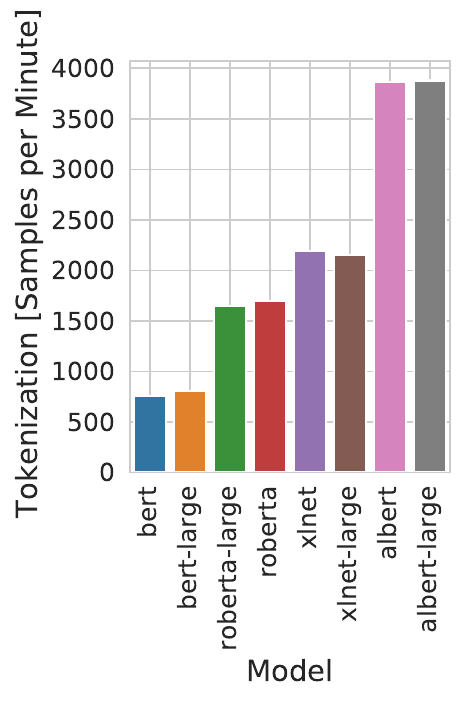}\includegraphics[width=0.4\textwidth]{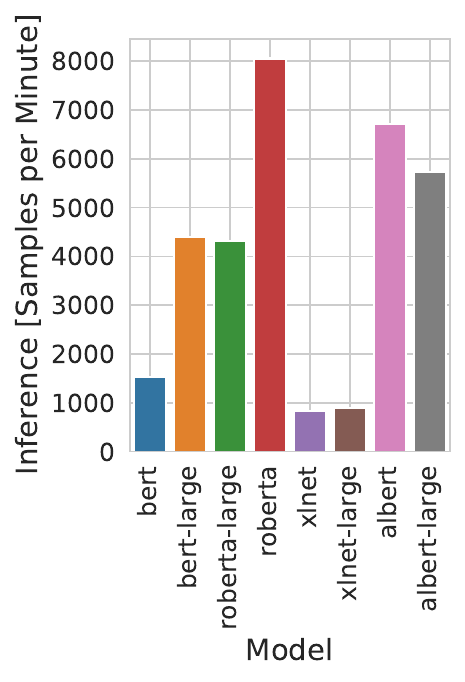}
    \caption{Language model performance (tokenization): samples per minute}
    \label{fig:tokperff1} 
\end{figure}

Language models were tested for performance using LightGBM model, batch size 32, time averaged over 4 batches. Time measurements obtained for total per batch processing time, tokenization time and model inference time. We compute performance in samples per minute terms. We observe that BERT and ROBERTA are the fastest models overall (see Figure \ref{fig:totalperff1}) with large variants of these models performing consistently slower up to 12 times difference (ALBERT vs ALBERT-large). We also see that speed of tokenization is very different with a noticable trend towards more recent and advanced variations of BERT (e.g. BERT has the slowest tokenization engine and ALBERT-LARGE is the fastest with 8 times difference in speed) - see Figure \ref{fig:tokperff1}. The speed of inference for BERT model is among three slowest models, with ROBERTA as the fastest models for inference (model run when tokenization is already performed over text lines).

In this study, we evaluated the performance of various language models using LightGBM with a consistent batch size of 32. The performance metrics were time-averaged across four batches, including total batch processing time, tokenization time, and model inference time. To express the efficiency of the models, we calculated their throughput in terms of samples per minute.

The results, illustrated in Figure \ref{fig:totalperff1}, indicate that the BERT and ROBERTA models shows the highest overall speed. In comparison, their larger counterparts, such as ALBERT and ALBERT-large, demonstrate a significant decrease in performance, with the latter being up to twelve times slower than its base model.

A notable variance in tokenization speeds was also observed, as depicted in Figure \ref{fig:tokperff1}. The trend suggests that more recently developed variations of the BERT model have improved tokenization efficiency, with BERT displaying the slowest tokenization process and ALBERT-LARGE being the fastest – the latter is eight times swifter than BERT.

During the model inference phase, where the models generate predictions post-tokenization, the BERT model's inference speed is amongst the lowest. The ROBERTA model stands out as the fastest during this stage, confirming its superiority in processing speed over the other models evaluated.

%

\subsection{General Comparison for Area: USA, Description only}

In this scenario we try to evaluate models on the task of traffic incident severity classification using description field only. We compare performance of language models by extracting NLP-features from accident descriptions: the NLP features are extracted using LLM directly from the 'description' field in the incident reports. The evaluation metric chosen is the F1-score, which balances precision and recall and is particularly useful for imbalanced classification problems (traffic incident severity classes are often imballanced).
\begin{figure*}[h]
    \centering
    \includegraphics[width=0.75\textwidth]{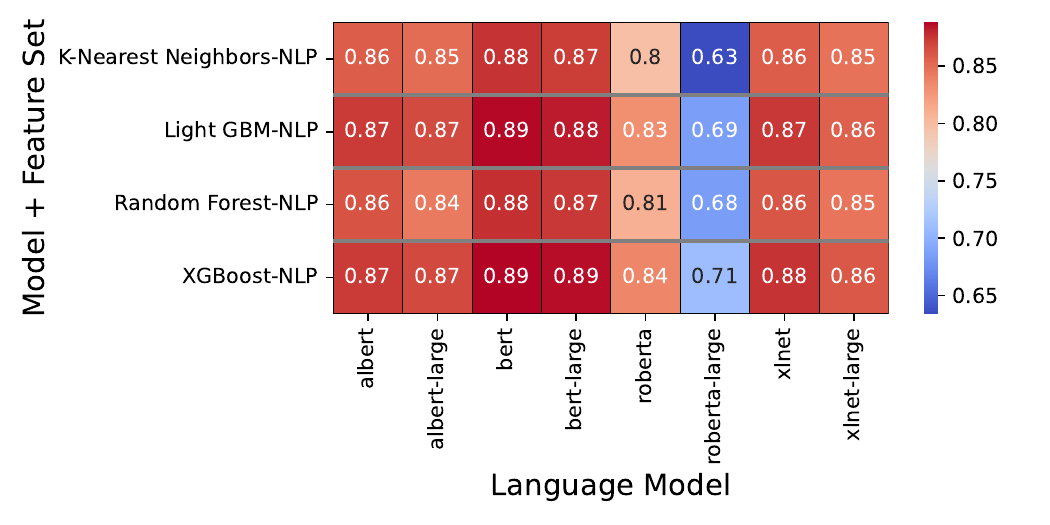}
    \caption{General comparison of Average F1 Score for traffic accident severity classification on USA data set using incident description only}
    \label{fig:descusa} 
\end{figure*}

There are several findings related to this description-only scenario (see Figure \ref{fig:descusa}):
\begin{itemize}
\item Effectiveness of BERT: Among the LLMs, BERT and its large variant outperform all other models in feature extraction relevant to incident severity classification. 
\item Random Forest and XGBoost: Both of these algorithms are ensemble methods that combine multiple weak learners to make a strong learner. These machine learning models are the most effective in utilizing the LLM-extracted features for severity classification. 
\item NLP Features extracted from the incident description field can be nearly as affective as Report-Only features from USA data set: this suggests that the unstructured text in the description contains nuances or contextual information that the tabular form does not capture. The results indicate that the textual description in the incident reports contains valuable information, necessary to detecrmine the accident severity. This makes sense, as the incident description might include details about how the accident occurred, how many vehicles were involved and other, even though in unstructured form.
\end{itemize}

The findings suggest that more focus of reporting authorities should be placed on extracting high-quality incident descriptions, as these are found to be very informative for the task at hand.  Also, it means that LLM models with ML pipeline can be utilized right away (e.g. in mobile application) once the accident report is received and provide a high degree of classification accuracy.
It would be interesting to explore why exactly BERT models are so effective in this scenario compared to other LLMs.

\subsection{Use of PCA for Dimensionality Reduction}

Given the high dimensionality of the feature set extracted from traffic incident reports using LLMs (e.g. 768 dimensions for each report for bert-large model), traditional machine learning models may face challenges in handling the data efficiently. The model training process can be time-consuming, high dimensionality can also lead to overfitting or poor generalization of the data. 

To mitigate these issues, we used fast machine learning models, like XGBoost and Decision Tree Regression model, are known for their efficiency and scalability in handling high-dimensional data. these models can process a large amount of data relatively quickly, enabling rapid model training and prediction, which is crucial for timely traffic incident response.

Furthermore, we used Principal Component Analysis (PCA) for dimensionality reduction. PCA is a popular technique that transforms the original variables into a new set of variables, which are linear combinations of the original variables and are orthogonal to each other, ensuring no redundant information. The transformation retains most of the variance in the data using fewer components, thus reducing the data's dimensionality.

\begin{figure}[h]
    \centering
    \includegraphics[width=0.4\columnwidth]{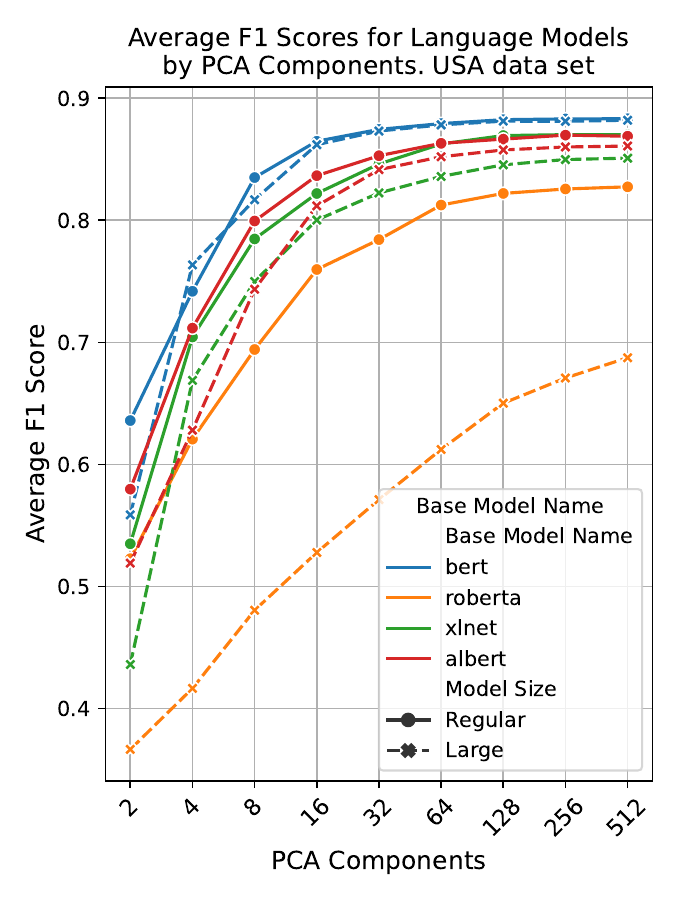}
    \caption{General comparison of Average F1 Score for traffic accident severity classification on USA data set using incident description only}
    \label{fig:scalef1} 
\end{figure}

Dimensionality reduction was applied to language feature vectors using Principal Component Analysis, as illustrated in Figure \ref{fig:scalef1}. The F1 scores from traffic accident severity classification were compared between language models, utilizing LightGBM as the machine learning model on a dataset from the USA. The findings suggest that base models works well with dimensionality reduction, showing a minimal difference in performance between 64 and 768 components. Conversely, large models demonstrate a distinctly slower progression, indicating a requirement for a larger number of components (e.g. albert vs albert-large), or the use of the full feature vector, to achieve sufficient results. The worst performance recorded for roberta-large, which for unknown reason has very slow progression in performance over number of components, which hints at requirement for full feature vector.

\section{Conclusion}\label{conclusions}

In the present study, we propose an innovative study on the efficiency of large language models for predicting traffic incident severity. We utilize the unstructured data representation across various traffic incident reports. By combining the extracted features from these reports with features extracted by large language models, our methodology shows enhancement in prediction accuracy. 

Our findings show that the integration of LLM-derived features with those from incident reports leads to a notable improvement in the performance of all tested machine learning models on USA dataset. Specifically, RandomForest and XGBoost exhibit the highest classification accuracy, signifying the potential of combining traditional machine learning with advanced NLP techniques for more accurate and efficient incident management.

The study illustrates LLMs' potential in enhancing TIMS through improved traffic incident severity classification. By applying advanced language models in conjunction with traditional machine learning, we can better understand and predict the outcomes of traffic incidents. Our approach represents a paradigm shift from relying solely on structured accident reports with pre-defined set of parameter variations to integrating the rich, contextual information embedded within the narrative of traffic accident reports and utilizing completely textual accident representation. Additionally, our evaluation revealed the trade-off between performance and computational cost across LLMs, providing a valuable reference for choosing suitable models based on the application requirements.

\textbf{Comparison with Tabular Data}: When compared with traditional, tabular accident report features (report), the LLM-derived features sometimes show either superior or acceptable performance, 
\textbf{Best Models}: For tasks like traffic incident severity classification, using advanced NLP models like BERT, coupled with powerful ML algorithms like Random Forest or XGBoost, is likely to yield the best results.
\textbf{Mitigating Feature Engineering}: Instead of relying solely on tabular data from accident reports, incorporating text-based features could provide a more holistic view and improve classification performance.

\textbf{Limitations of this work:} Our current approach has been primarily tested and validated on publicly available CTADS dataset from the United States. Nevertheless, there is a potential in applying this approach to traffic incident data from other countries.

\textbf{Future works:} By utilizing the flexibility of LLMs in interpreting incident reporst represented in the textual form, we can build traffic incident severity classification frameworks, which have the potential to be transferable between various accident reporting systems and locations. Our future research will also focus on performing a joint prediction of traffic accident severity, impact and duration. LLM models can be fine-tuned for specific tasks, including the interpretation of technical terms, abbreviations, and numbers. However, out-of-the-box, neither is specifically designed to perform well in handling highly abbreviated or numerical short strings present in accident reports. For such specialized tasks, a domain-specific model fine-tuned on a large corpus of accident reports would likely perform better.



{
\section*{Acknowledgment}
This work has been done as part of the ARC Linkage Project LP180100114.
}

\newpage


\bibliographystyle{elsarticle-num}
\bibliography{trb_template}

\end{document}